%% file: main.tex
\documentclass[sigconf,screen]{acmart}

\PassOptionsToPackage{table}{xcolor}%\usepackage[table]{xcolor} 
\usepackage{colortbl}
\usepackage{url}
\usepackage{adjustbox}
\usepackage{algorithm}
\usepackage{algpseudocode}
\usepackage{multirow}
\usepackage{tikz}
\usepackage{amsmath}
\usepackage{xspace}
\usepackage{enumitem}
\usepackage{booktabs}
\usepackage{bbding}
\usepackage{subfig}
\usepackage{stfloats}
\usepackage{indentfirst}
\usepackage{makecell}
\usepackage[font=small]{caption}
\usepackage{tcolorbox}
\usepackage{lipsum}
\usepackage{minted}
\usepackage{graphicx}
\usepackage[normalem]{ulem}
\usepackage{minted} 
\usepackage{multicol}
\usepackage{enumitem}
\usepackage{graphicx}
\usepackage{subcaption}
\usepackage{array}
\usepackage{ragged2e}
\usepackage{tabularx}
\usepackage[table]{xcolor} 
\usepackage{booktabs} 
\usepackage{caption}
\usepackage{soul}
\usepackage[]{collab}
\tcbuselibrary{listingsutf8}
\tcbuselibrary{skins}
\tcbset{
  mymintedbox/.style={
    colback=gray!20, % 背景色
    colframe=gray!50, % 边框色
    boxrule=0.5pt, % 边框宽度
    arc=3pt, % 圆角
    boxsep=0pt,
    left=2pt,
    right=2pt,
    top=1pt,
    bottom=1pt,
    fontupper=\ttfamily\footnotesize, % 设置字体为等宽字体和小号字体
    listing options={breaklines=true, breakautoindent=true, breakindent=2em} % 自动换行
  }
}

\author{Jia Li}
\authornote{Both authors contributed equally to this research.}
\orcid{0009-0007-5557-4916}
\email{linsayli@link.cuhk.edu.hk}
\affiliation{%
  \institution{The Chinese University of Hong Kong}
  \city{Hong Kong}
  \country{Hong Kong}
}

\author{Wenyuan Ma}
\authornotemark[1]
\orcid{0009-0003-2341-1945}
\email{willwyma@tencent.com}
\affiliation{%
  \institution{Tencent Inc.}
  \city{Shenzhen}
  \country{China}
}

\author{Ting Peng}
\orcid{0009-0003-6970-0857}
\email{sakurapeng@tencent.com}
\affiliation{%
  \institution{Tencent Inc.}
  \city{Shenzhen}
  \country{China}
}

\author{Haibin Zheng}
\orcid{0009-0007-5080-3070}
\email{mattzheng@tencent.com}
\affiliation{%
  \institution{Tencent Inc.}
  \city{Shenzhen}
  \country{China}
}

\author{Yuetang Deng}
\authornote{Corresponding author.}
\orcid{0009-0003-7060-4109}
\email{yuetangdeng@tencent.com}
\affiliation{%
  \institution{Tencent Inc.}
  \city{Shenzhen}
  \country{China}
}

\AtBeginDocument{%
  }

\ccsdesc[500]{Software and its engineering~Software testing and debugging}
\ccsdesc[500]{Computing methodologies~Natural language processing~Information extraction}

\keywords{Automated Crash Diagnosis, Multi-Agent Systems, Large Language Models, Fault Localization, Mobile Applications}

\begin{document}

\title{Holmes: Multimodal Agentic Diagnosis for Mixed-Language Mobile Crashes at Industrial Scale}

\input{sections/00-abstract}
\maketitle
\input{sections/01-introduction}
\input{sections/04-technique}

\input{sections/05-evaluation}

\input{sections/06-conclusion}

\enlargethispage{-13pc}
%\balance
\bibliographystyle{ACM-Reference-Format}
\bibliography{main}

\end{document}

%% file: sections/00-abstract.tex
\begin{abstract}
Diagnosing mobile crashes in ultra-large-scale industrial applications is a formidable challenge due to the sheer volume of code, the complexity of mixed-language environments, and the inability to reproduce failures locally. Traditional static analysis struggles with scalability, while existing LLM-based agents often rely on reproducible environments unavailable in post-mortem scenarios. We present Holmes, a multi-agent system that automates root cause analysis by synthesizing multimodal runtime signals---stack traces, logs, and thread states---to reconstruct failure contexts without reproduction. Holmes introduces a hierarchical Retrieve-Explore-Reason architecture that leverages low-level artifacts (e.g., registers, assembly) to bridge the semantic gap between open-source business logic and closed-source system frameworks. By dynamically compressing the search space using runtime clues, Holmes precisely navigates 70-million-line codebases to identify non-local defects. Evaluated on real-world crashes from WeChat, Holmes achieves 87.6\% accuracy in function-level fault localization and reduces average investigation time by over 98\% (to $\sim$77 seconds), demonstrating its effectiveness in transforming labor-intensive debugging into an efficient verification workflow.
\end{abstract}

%% file: sections/01-introduction.tex
\section{Introduction}

Modern mobile applications have evolved into complex ecosystems serving billions of users. In WeChat, the client generates millions of crash reports daily. While automated services cluster crashes based on stack trace similarity, they lack automated root cause analysis (RCA) and actionable fix suggestions. Developers must manually map runtime symptoms to static code defects, a daunting task in ultra-large-scale industrial repositories. Bridging the gap between runtime data and source code requires deep domain knowledge and hours of investigation. Internal data shows that complex crash clusters typically require 2 to 3 hours of ticket handling time to reach an actionable diagnosis (Time-to-Insight), involving labor-intensive correlation of multimodal clues across stacks, logs, thread states, user trajectories, and source code.

In large industrial repositories, the gap between dynamic symptoms and static logic widens. Traditional static analysis \cite{Infer} is often impractical at the scale of 70 million lines of code due to the computational cost of global inter-procedural analysis. Recent LLM-based approaches \cite{Wu2025} using pre-built call graphs face high maintenance costs and limited scalability in systems with over 6 million functions. Furthermore, state-of-the-art software engineering agents \cite{SWEagent, Openhands, Agentless} typically rely on reproducible environments and failure-inducing tests, which are unavailable for post-mortem diagnosis of mobile crashes. Given the diversity of user environments and privacy constraints that prevent data access, local sandboxes cannot replicate the transient states of millions of devices. This necessitates a post-mortem reasoning paradigm capable of navigating massive repositories to localize root causes using only sparse, read-only artifacts, without requiring a reproducible environment.

Despite the promising reasoning capabilities of LLMs, automated diagnosis in ultra-large-scale industrial environments faces significant hurdles. First, existing methods typically rely on single-modal analysis, focusing either on logs~\cite{LogLLM, LogAnomaly, LogPPT, FaithLog} or code~\cite{Wu2025, LocAgent, RepoGraph, CodePlan, ReBucket}. This isolation fails to synthesize the spatial, temporal, and concurrency clues essential for complex crashes. Second, current techniques struggle with the semantic gap in mixed-source environments. Most approaches assume full source visibility and hit dead ends when execution flows into closed-source binaries, unable to trace logic across system boundaries. Third, navigating massive search spaces for non-local root causes is computationally prohibitive; standard retrieval struggles to locate distant defects without overwhelming context windows. Finally, the lack of low-level runtime signals (e.g., registers, memory snapshots) limits hypothesis verification in non-reproducible, post-mortem scenarios.

In this paper, we propose Holmes, a multi-agent system that formulates crash diagnosis as an agentic reasoning task. To overcome single-modal and scalability limitations, Holmes performs joint reasoning over the full spectrum of dynamic runtime signals available in system dumps---stack traces, time-ordered log events, and concurrent thread states. This multimodal synthesis not only reconstructs the full failure context but also aggressively compresses the code search space, enabling precise navigation to non-local root causes within 70 million lines of code. Crucially, Holmes integrates low-level artifacts like registers and assembly to bridge the semantic gap between closed-source system frameworks and open-source business logic, enabling hypothesis verification across these boundaries. By adopting a hierarchical Retrieve-Explore-Reason architecture, Holmes synthesizes these heterogeneous clues to deliver industrial-grade localization, explanation, and fix suggestions in approximately one minute.

We evaluated Holmes using a dataset of 73 real-world crash reports sampled from the WeChat iOS production environment. Senior developers defined a taxonomy and annotated the dataset with ground-truth fault locations and root causes. On this complex dataset, Holmes achieves 87.6\% pass@1 accuracy for function-level fault localization and 65.7\% for root-cause identification, with a mean latency of 168.5 seconds (compared to 77 seconds average in large-scale production). By synthesizing multimodal evidence to navigate millions of functions, Holmes reduces investigation time by over 98\%, shifting the workflow from manual investigation to efficient verification.

\noindent \textbf{Contributions.} This paper makes the following technical contributions:
\begin{itemize}[noitemsep, topsep=0pt, partopsep=0pt, leftmargin=4mm]
\item \textbf{Multimodal Agentic Diagnosis Framework.} We propose Holmes, a multi-agent system that formulates crash diagnosis as a collaborative reasoning task. By synthesizing heterogeneous signals from stacks, logs, threads, and code, it overcomes the limitations of single-modal approaches and reconstructs complex failure contexts.
\item \textbf{Hybrid-Source Reasoning with Low-Level Artifacts.} We introduce a method to bridge the semantic gap in mixed-source environments by integrating registers, memory snapshots, and assembly code. This enables seamless reasoning across the boundary between open-source business logic and closed-source system frameworks.
\item \textbf{Scalable Navigation via Dynamic Compression.} We present a retrieval-augmented strategy that leverages dynamic runtime signals to aggressively compress the search space. This allows precise navigation to non-local root causes within a 70-million-line repository within minutes, without relying on expensive global static analysis.
\item \textbf{Industrial-Scale Evaluation.} We evaluate Holmes on real-world crashes from WeChat. Results show it achieves 87.6\% accuracy in function-level localization and reduces investigation time by over 98\% (to $\sim$77 seconds), demonstrating its effectiveness and efficiency in a high-volume production environment.
\end{itemize}

%% file: sections/04-technique.tex
\section{Methodology}

\begin{figure*}
    \centering
    \includegraphics[width=0.95\textwidth]{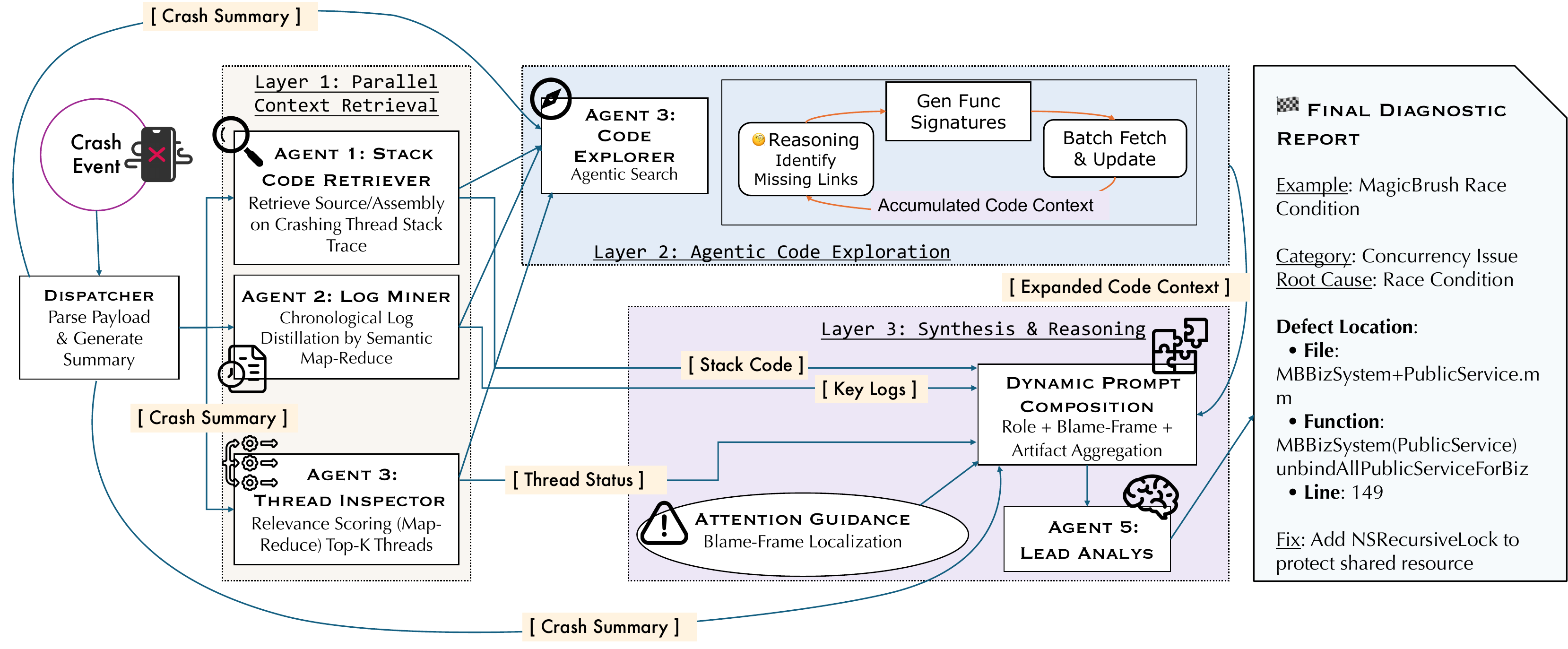}
    \caption{System architecture of the Holmes framework. It illustrates the tripartite workflow: parallel context retrieval via specialized agents (Layer 1), iterative logic navigation across massive codebases (Layer 2), and multimodal evidence synthesis for final diagnosis (Layer 3).}
    \label{fig:arch}
\end{figure*}

Holmes is designed for industrial crash diagnosis in mixed-language codebases (C/C++/ObjC/Swift), leveraging multi-modal artifacts including stacks, logs, code, and rich metadata. It adopts a hierarchical Retrieve-Explore-Reason architecture (Figure~\ref{fig:arch}) with three layers: (1) \textbf{Parallel Context Retrieval} for evidence collection; (2) \textbf{Agentic Code Exploration} for deep root cause search; and (3) \textbf{Synthesis \& Reasoning} for final diagnosis. This design ensures focused data extraction before global synthesis.

\subsection{Layer 1: Parallel Context Retrieval}

\noindent\textbf{Crash Summary}
In the production environment, crash reports are captured on user devices and uploaded to a central server. The diagnosis workflow is initiated by the \textbf{Dispatcher}, a core component of the Holmes framework responsible for orchestrating the multi-agent collaboration. Upon receiving a raw crash report, the Dispatcher first parses the raw data to extract three primary artifacts: the crashing thread's stack trace, the full log stream, and relevant source/assembly code. Remaining key fields (e.g., exception type, user click paths, register states, memory stats) are normalized into \textbf{metadata} (Table~\ref{tab:metadata}). This metadata serves as stable \textbf{grounding signals} injected into downstream agents to constrain and calibrate subsequent evidence extraction.

\begin{table*}[t]
\centering
\small
\setlength{\tabcolsep}{2pt}
\renewcommand{\arraystretch}{1.05}
\caption{Metadata fields extracted from crash reports.}
\label{tab:metadata}
{\arrayrulecolor{gray!30}\rowcolors{2}{gray!8}{white}%
\begin{tabularx}{\textwidth}{@{}>{\raggedright\arraybackslash}p{0.15\textwidth} >{\raggedright\arraybackslash}p{0.20\textwidth} >{\raggedright\arraybackslash}X @{}}
\rowcolor{gray!20}
\textbf{Field} & \textbf{Explanation} & \textbf{Value} \\
\midrule
Basic Info & Exception, OS, Device info & \texttt{Exception: EXC\_BAD\_ACCESS}; \texttt{Process: WeChat (9.8.241)}; \texttt{Device: iPhone12,1}; \texttt{OS: iOS 16.3.1} \\
User Click Path & View controller trajectory & \texttt{MultiSelectContactsVC} $\rightarrow$ \texttt{MinimizeVC} $\rightarrow$ \texttt{NewMainFrameVC} \\
Crashed Thread & Top frame of crashing thread & \texttt{Thread 0: com.apple.main-thread}; \texttt{Top: \_\_CFStringAppendBytes $\rightarrow$ ...} \\
Registers & Key CPU registers (pc, lr, sp) & \texttt{pc: 0x1c5565d60}; \texttt{lr: 0x1910dc01c5533c64}; \texttt{sp: 0x16f9afdf0} \\
Memory Snapshots & Relevant variable values & \texttt{x1 (0x117aba2ee): migration\_info}; \\
VM Summary & Virtual and resident memory & \texttt{Virtual: 9.7G / Resident: 391.4M}; \texttt{Stack: 94.2M} \\
\bottomrule
\end{tabularx}}
\arrayrulecolor{black}
\end{table*}

\subsubsection{Agent 1: The Stack Code Retriever (Focus: Static Stack).}
This agent provides the baseline code context for the crashing thread by taking the crashing thread stack trace as input and outputting initial stack code snippets (source code or assembly). Its primary role is to collect the code executed at the time of the crash before deep exploration, performing retrieval only without conducting reasoning. To implement this, the agent extracts code for all functions in the crashing thread's stack trace using a dual-mode retrieval policy based on frame type: it retrieves source code (e.g., MessageMgr.mm) for internal business logic to support understanding, and assembly code (e.g., libobjc.A.dylib) for system frameworks when source code is unavailable, ensuring complete visibility into the execution path across system boundaries.

\subsubsection{Agent 2: The Log Miner (Focus: Runtime Logs).} 
This agent identifies time-ordered events relevant to the crash by processing the Crash Summary and Raw Log Stream to output filtered key log snippets. Unlike naive approaches that simply truncate logs to a fixed size (e.g., the last 200 lines), which risks missing early precursors, we capture a broader 1-minute temporal window and employ a semantic Map-Reduce strategy to distill relevant signals. The implementation begins by building a Crash Summary (including crash metadata, crashing thread context, and key stack symbols) as an anchor. To mitigate the lost-in-the-middle phenomenon, the system splits the 1-minute log window into 8k-token chunks and processes them with four parallel workers. These workers are explicitly instructed to retain only log lines that share causal dependencies (e.g., resource allocation, state transitions) with the entities identified in the Crash Summary. Finally, these filtered segments are aggregated into a single timeline of relevant events.

\subsubsection{Agent 3: The Thread Inspector (Focus: Concurrency State).} 
The Thread Inspector detects cross-thread interference, such as deadlocks and race conditions, by analyzing the Crash Summary and Full Thread Dump to output the top 2 high-relevance thread stacks. Since analyzing all threads uniformly introduces noise and exceeds context limits, we formulate thread analysis as a relevance ranking problem using a Map-Reduce strategy. The system groups the full thread dump into chunks (approx. 8k tokens) with atomic thread boundaries. Six parallel workers assign relevance scores to each thread against the Crash Summary in the Map Phase, and the results are merged and sorted based on these scores in the Reduce Phase to retain the most relevant threads.

\subsection{Layer 2: Agentic Code Exploration}

A crash stack trace often reflects the observed failure, while the underlying cause may reside in asynchronous callbacks or in earlier execution paths.

\textbf{Scalability Challenge.} In a 70M LOC repository (6M+ functions), maintaining a global static Call Graph (CG) is impractical. Internal benchmarks show generating an inter-procedural CG requires >12 hours and >500GB memory, blocking CI integration. Similarly, RAG approaches~\cite{GraphRAG, RAG}  face scalability hurdles, with index construction taking >7 days and struggling with updates. Furthermore, dynamic dispatch in ObjC/C++ renders static graphs inherently incomplete.

\textbf{Agentic Search Paradigm.} Instead of pre-built graphs, Holmes adopts an agentic search paradigm supported by a lightweight \textbf{function-to-file-path index} for on-demand navigation. Initial indexing takes $\sim$3.3 hours (6GB footprint), with incremental updates taking only $\sim$10 minutes.

Building on this foundation, the layer consumes the Crash Summary together with the Initial Context (Stack Code Snippets from Agent 1, Key Logs from Agent 2, and Relevant Threads from Agent 3) and incrementally produces an Expanded Code Context that accumulates the retrieved function implementations. The core mechanism is a loop (maximum 3 iterations) that progressively expands the context: (1) \textbf{Context Assembly}: Formats accumulated evidence; (2) \textbf{Reasoning \& Decision}: The LLM identifies missing logical links and outputs target function signatures; (3) \textbf{Batch Retrieval}: Fetches implementations in parallel; and (4) \textbf{State Update}: Accumulates new snippets. The loop terminates when no new targets are found.

\subsection{Layer 3: Synthesis \& Reasoning}
Acting as the lead analyst, this layer is responsible for global evidence synthesis and final report generation. It takes the Crash Summary, Expanded Code Context, Stack Code Snippets, Key Logs, Relevant Threads, and the Blame-Frame Prompt as input, and produces a Final Diagnostic Report containing the root cause, defect localization, evidence chain, and fix suggestions. The workflow consists of four steps: (1) Attention Guidance (Blame-Frame Localization), which predicts a single suspicious \texttt{file:line} anchor to force reasoning to stay anchored; (2) Dynamic Prompt Composition, which aggregates refined outputs and conditionally appends specialized sections; (3) Heuristic Rule Injection, guided by internal statistics showing that memory errors account for over 50\% of crashes, injects a minimal set of domain rules (e.g., checking low addresses to distinguish null pointers from wild pointers) to resolve specific ambiguities; and (4) Final Reasoning, which, instead of relying on the LLM's inherent capabilities alone, leverages the structured, semantically aligned evidence prepared by the previous layers to enable rigorous cross-modal consistency checks and generate the final report.

%% file: sections/05-evaluation.tex
\section{Experiment}

To validate Holmes' effectiveness in an industrial setting, we conducted a comprehensive evaluation using real-world crash reports labeled by senior developers.

\noindent\textbf{Crash Taxonomy Definition}
To ensure a rigorous and standardized evaluation, we established a taxonomy of crash categories and root causes. This taxonomy is derived from the Common Weakness Enumeration (CWE) framework and refined based on the statistical distribution of failure modes in our production environment. These categories are detailed in Table~\ref{tab:taxonomy}.

\begin{table*}[t]
\centering
\small
\setlength{\tabcolsep}{4pt}
\renewcommand{\arraystretch}{1.1}
\caption{Detailed crash taxonomy and root cause definitions.}
\label{tab:taxonomy}
{\arrayrulecolor{gray!30}\rowcolors{2}{gray!8}{white}%
\begin{tabularx}{\textwidth}{@{}>{\raggedright\arraybackslash}p{0.15\textwidth} >{\raggedright\arraybackslash}p{0.25\textwidth} >{\raggedright\arraybackslash}X@{}}
\rowcolor{gray!20}
\textbf{Category} & \textbf{Root Cause} & \textbf{Core Description} \\
\midrule
\textbf{Memory Error} & Null Pointer Dereference & Attempting to read or write memory pointed to by a NULL or nullptr pointer. \\
 & Garbage Pointer & Accessing memory that has been freed (dangling pointer) or uninitialized (wild pointer). \\
 & Invalid Free & Attempting to free a pointer not allocated by malloc/new or already freed. \\
 & Buffer Overflow & Writing data beyond the boundaries of an array or buffer. \\
 & Stack Overflow & Infinite recursion or allocating excessively large local variables on the stack. \\
\midrule
\textbf{Logic Error} & Division by Zero & Denominator is zero in a mathematical operation. \\
 & Unhandled Exception & Program throws an exception without any try-catch block to handle it. \\
 & Assertion Failure & Condition of assert() macro is false, causing program termination. \\
 & Unrecognized Selector & Dynamic language (e.g., ObjC) calls an unimplemented method on an object. \\
 & Type Conversion Failure & Runtime type casting check fails (e.g., in Swift). \\
 & Null Parameter & Passing a null value to a function that expects a non-null argument. \\
\midrule
\textbf{Concurrency} & Race Condition & Multiple threads access shared data concurrently without synchronization. \\
 & Deadlock & Two or more threads waiting for each other to release resources. \\
\midrule
\textbf{External \& Env} & Improper API Usage & Violating usage contracts when calling third-party libraries or system APIs. \\
 & Corrupted Input & Crash during parsing of malformed or corrupted files/data streams. \\
\midrule
\textbf{Resource Mgmt} & Main Thread Violation & Executing UI operations on a non-main thread. \\
 & Object Lifecycle & Improper timing of object creation/destruction, or internal state inconsistency. \\
 & Resource Leak & Leaking file descriptors or contexts. \\
 & Disk Space Exhaustion & Storage operations fail due to insufficient space. \\
 & Out of Memory & Allocation failure due to insufficient memory. \\
\bottomrule
\end{tabularx}}
\arrayrulecolor{black}
\end{table*}

\subsection{Dataset \& Ground Truth}

We collected a dataset of 73 crash reports from the WeChat iOS client.

\textbf{Sampling:} Given the labor-intensive nature of manual root cause analysis, we constructed a focused dataset of 73 cases. We performed stratified sampling from the top 500 most frequent crash clusters. This type-based stratified sampling approach preserves the true distribution of the most frequent failure modes in the production environment.

\textbf{Diversity:} The dataset spans multiple critical chains covering diverse modules ranging from low-level infrastructure (network, storage, kernel) to high-level UI and business logic (multimedia, mini-programs). Based on the 73 valid cases used for calculation, the failure modes are distributed as follows: Memory Errors (53.4\%), Logic Errors (23.3\%), External \& Environmental Issues (13.7\%), Resource Management Issues (4.1\%), and Concurrency Issues (5.5\%). This distribution reflects the high prevalence of memory safety issues in large-scale codebases using C++ and Objective-C.

\textbf{Labeling:} To prevent bias, the labeling process was conducted independently of and prior to the Holmes diagnosis. Each report was manually analyzed by two senior developers using a cross-validation process to establish the Ground Truth for the category, root cause, and defect location (file/function/line). All three localization labels refer to the defect-causing (most attributable) location as identified by developers (i.e., the file/function/line where the defect was introduced or that is causally responsible for the crash), rather than the patch location. Disagreements were resolved through discussion to ensure high-quality labels.

\subsection{Accuracy Results}

We define Accuracy as the percentage of cases where the AI-generated output matches the ground truth. We evaluated accuracy across five dimensions. The results on our dataset ($N=73$) are shown in Table~\ref{tab:accuracy}.

\begin{table*}[t]
\centering
\begin{minipage}{0.48\textwidth}
\centering
\small
\setlength{\tabcolsep}{3pt}
\renewcommand{\arraystretch}{1.08}
\caption{Evaluation of Holmes accuracy across five dimensions ($N=73$).}
\label{tab:accuracy}
{\arrayrulecolor{gray!30}\rowcolors{2}{gray!8}{white}%
\begin{tabularx}{\linewidth}{@{}>{\raggedright\arraybackslash}X c@{}}
\rowcolor{gray!20}
\textbf{Metric (Definition)} & \textbf{Accuracy} \\
\midrule
Category Accuracy (Correct classification of crash type) & 79.4\% \\
Root Cause Accuracy (Identification of fundamental reason) & 65.7\% \\
Function Accuracy (Identifying exact defect function) & 87.6\% \\
File Accuracy (Identifying correct source file) & 90.4\% \\
Line Accuracy (Pinpointing defect-causing source line) & 87.6\% \\
\bottomrule
\end{tabularx}}
\arrayrulecolor{black}
\end{minipage}
\hfill
\begin{minipage}{0.48\textwidth}
\centering
\small
\setlength{\tabcolsep}{3pt}
\renewcommand{\arraystretch}{1.08}
\caption{Ablation study results comparing Holmes with variants and baseline.}
\label{tab:ablation}
{\arrayrulecolor{gray!30}\rowcolors{2}{gray!8}{white}%
\begin{tabularx}{\linewidth}{@{}>{\raggedright\arraybackslash}p{0.35\linewidth}>{\centering\arraybackslash}p{0.3\linewidth}>{\centering\arraybackslash}p{0.35\linewidth}@{}}
\rowcolor{gray!20}
\textbf{Model / Variant} & \textbf{Function Accuracy} & \textbf{Root Cause Accuracy} \\
\midrule
\rowcolor{gray!15}
\textbf{Holmes (Full)} & \textbf{87.6\% (64/73)} & \textbf{65.7\% (48/73)} \\
\textit{w/o} Low-level Artifacts & 84.2\% (61/73) & 57.5\% (42/73) \\
\textit{w/o} Attention Guidance & 81.5\% (59/73) & 56.2\% (41/73) \\
\textit{w/o} Thread Agent & 78.1\% (57/73) & 63.0\% (46/73) \\
\textit{w/o} Log Agent & 83.6\% (61/73) & 52.1\% (38/73) \\
\textit{w/o} Code Explorer & 68.5\% (50/73) & 60.3\% (44/73) \\
Vanilla DeepSeek-V3.1 & 42.5\% (31/73) & 35.6\% (26/73) \\
\bottomrule
\end{tabularx}}
\arrayrulecolor{black}
\end{minipage}
\end{table*}

Holmes demonstrates robust fault localization, the most labor-intensive debugging phase. High \textbf{Function (87.6\%)} and \textbf{File (90.4\%) Accuracy} confirm that the retrieval system effectively narrows the search space to the correct code region. Comparable \textbf{Line Accuracy} indicates precise defect pinpointing once the region is locked. \textbf{Category Accuracy (79.4\%)} remains strong, effectively distinguishing defect types despite occasional confusion between memory and logic errors. While \textbf{Root Cause Accuracy (65.7\%)} is lower due to semantic ambiguities (e.g., null vs. garbage pointers), high localization accuracy ensures developers are guided to the correct verification region. Regarding efficiency, the average diagnostic latency of 168.5 seconds (approx. 2.8 minutes) on this complex dataset is higher than the production average (77s) due to deeper exploration, yet still represents an order-of-magnitude improvement over manual debugging.

\subsubsection{Per-Category Accuracy Breakdown}

To provide a finer-grained view of Holmes' diagnostic capabilities across the taxonomy defined in Table~\ref{tab:taxonomy}, we report per-category accuracy in Table~\ref{tab:per_category} and per-root-cause accuracy in Table~\ref{tab:per_rootcause}.

\begin{table*}[t]
\centering
\small
\setlength{\tabcolsep}{4pt}
\renewcommand{\arraystretch}{1.1}
\caption{Per-category accuracy breakdown ($N=73$). $n$: number of cases per category.}
\label{tab:per_category}
{\arrayrulecolor{gray!30}\rowcolors{2}{gray!8}{white}%
\begin{tabular}{l c c c c c c}
\rowcolor{gray!20}
\textbf{Category} & \textbf{$n$} & \textbf{Cat. Acc.} & \textbf{RC Acc.} & \textbf{Func. Acc.} & \textbf{File Acc.} & \textbf{Line Acc.} \\
\midrule
Memory Error     & 39 & 97.4\% & 76.9\% & 94.9\% & 94.9\% & 89.7\% \\
Logic Error      & 17 & 58.8\% & 47.1\% & 82.4\% & 82.4\% & 82.4\% \\
External \& Env  & 10 & 30.0\% & 30.0\% & 60.0\% & 80.0\% & 80.0\% \\
Concurrency      &  4 & 100\%  & 100\%  & 100\%  & 100\%  & 100\%  \\
Resource Mgmt    &  3 & 100\%  & 100\%  & 100\%  & 100\%  & 100\%  \\
\bottomrule
\end{tabular}}
\arrayrulecolor{black}
\vspace{6pt}

\centering
\small
\setlength{\tabcolsep}{4pt}
\renewcommand{\arraystretch}{1.1}
\captionof{table}{Per-root-cause accuracy for root causes with $n \geq 2$.}
\label{tab:per_rootcause}
{\arrayrulecolor{gray!30}\rowcolors{2}{gray!8}{white}%
\begin{tabular}{l c c c | l c c c}
\rowcolor{gray!20}
\textbf{Root Cause} & \textbf{$n$} & \textbf{RC Acc.} & \textbf{Func. Acc.} & \textbf{Root Cause} & \textbf{$n$} & \textbf{RC Acc.} & \textbf{Func. Acc.} \\
\midrule
Null Pointer Dereference & 22 & 72.7\% & 95.5\% & Unrecognized Selector   &  3 & 100\%  & 66.7\% \\
Garbage Pointer       & 10 & 100\%  & 90.0\% & Unhandled Exception     &  2 & 50.0\% & 100\%  \\
Assertion Failure     &  9 & 33.3\% & 77.8\% & Corrupted Input         &  2 &  0.0\% &  0.0\% \\
Improper API Usage    &  8 & 37.5\% & 75.0\% & Invalid Free            &  2 &  0.0\% & 100\%  \\
Buffer Overflow       &  5 & 80.0\% & 100\%  & Null Parameter          &  2 &  0.0\% & 100\%  \\
Race Condition        &  4 & 100\%  & 100\%  & Main Thread Violation   &  2 & 100\%  & 100\%  \\
\bottomrule
\end{tabular}}
\arrayrulecolor{black}
\end{table*}

\textbf{Category-level analysis.}
Holmes achieves near-perfect classification for \textbf{Memory Errors} (97.4\% category accuracy), which constitute the majority of crashes (53.4\%). \textbf{Concurrency} and \textbf{Resource Management} categories, though small in sample size ($n=4$ and $n=3$), are diagnosed with 100\% accuracy across all metrics, demonstrating Holmes' strength in leveraging thread-level and runtime evidence. Performance is weaker on \textbf{Logic Errors} (58.8\%) and \textbf{External \& Environmental} issues (30.0\%), where the system tends to misclassify them as memory errors due to similar crash signatures (e.g., assertion failures manifesting as \texttt{SIGABRT}).

\textbf{Root-cause-level analysis.}
At the fine-grained root cause level (Table~\ref{tab:per_rootcause}), Holmes excels at structurally distinctive failure modes: \textbf{Garbage Pointer} (100\%), \textbf{Race Condition} (100\%), and \textbf{Unrecognized Selector} (100\%) are identified with perfect root cause accuracy. For \textbf{Null Pointer Dereference}---the most frequent root cause ($n=22$)---root cause accuracy is 72.7\% while function localization reaches 95.5\%, indicating that even when the specific root cause label is debated, the defect location is correctly identified.

The weakest root causes are \textbf{Assertion Failure} (33.3\%), \textbf{Improper API Usage} (37.5\%), \textbf{Corrupted Input} (0\%), \textbf{Invalid Free} (0\%), and \textbf{Null Parameter} (0\%). These share a common challenge: the crash symptom is indirect, requiring deeper semantic understanding of API contracts or data provenance that goes beyond what stack traces and code can directly reveal. Notably, even for these difficult cases, \textbf{function localization accuracy remains substantially higher} (66.7--100\%), confirming that Holmes successfully guides developers to the correct code region even when the precise root cause taxonomy label is contested.

\subsection{Comparative Analysis \& Ablation Study}

To justify the multi-agent design, we compare Holmes with a commonly used \textit{paste-the-stack-into-a-chatbot} baseline and conduct ablation studies (Table~\ref{tab:ablation}). We report \textbf{Pass@1} for function-level fault localization and root-cause identification based on our labeled dataset.

\textbf{Baseline:} We compared Holmes against \textbf{Vanilla DeepSeek-V3.1 (Zero-shot)}. The Vanilla baseline uses only raw crash stacks and exceptions, lacking repository retrieval, logs, or thread data. We excluded other state-of-the-art tools (e.g., SWE-agent~\cite{SWEagent}, RepoGraph~\cite{RepoGraph}) and standard RAG approaches~\cite{GraphRAG, RAG} because they are inapplicable to our industrial setting: they either require reproducible environments (unavailable in post-mortem scenarios) or cannot scale to 70-million-line codebases due to the prohibitive cost of global graph or vector index construction.

\textbf{Ablation Variants:} \textbf{w/o Log Agent} (no runtime logs); \textbf{w/o Thread Agent} (no cross-thread evidence); \textbf{w/o Code Explorer} (stack-visible code only); \textbf{w/o Attention Guidance} (no \texttt{file:line} anchor). \textbf{w/o Low-level Artifacts} (no registers/memory/assembly).

\textbf{Analysis:} \textbf{Holmes (Full)} balances high-recall retrieval with constrained reasoning. Multimodal data provides complementary causal views, while bounded exploration completes logical links missing from static snapshots, significantly reducing hallucinations. \textbf{w/o Low-level Artifacts:} Excluding low-level signals (registers, assembly) severs the link between high-level logic and binary execution states. This is particularly detrimental for mixed-source crashes (e.g., JNI, system frameworks), where register values are often the only clue to validate hypotheses, leading to a notable drop in root cause accuracy. \textbf{w/o Attention Guidance:} Performance drops confirm that blame-frame anchoring is a critical attention gate. Lacking \texttt{file:line} constraints leads to attention diffusion, prone to misattributing errors to irrelevant glue code. \textbf{w/o Thread Agent:} Missing thread data hinders the detection of concurrency interference (e.g., cross-thread race conditions, deadlock precursors). Since root causes are often non-local, removing this dimension causes the model to overfit to surface symptoms of the victim thread, thereby missing true concurrency accomplices. \textbf{w/o Log Agent:} Logs provide necessary runtime constraints (e.g., execution paths, I/O states, feature flags). Without logs, the model cannot reconstruct the scene, degrading reasoning to static guessing based on code, making it difficult to pinpoint specific root causes. \textbf{w/o Code Explorer:} A 19.1\% drop in localization accuracy confirms that on-demand exploration is vital for reconstructing causal chains. Relying solely on stack code leads to context truncation, unable to trace upstream logic errors. \textbf{Vanilla DeepSeek-V3.1:} Performs worst, revealing the limitations of weak-evidence reasoning. Lacking repository context and runtime states, LLMs can only make probabilistic guesses, unable to establish reliable symptom-defect mappings.

\textbf{Conclusion:} Experiments reveal that standalone LLM reasoning fails in industrial RCA due to the \textbf{lack of runtime context}. By dynamically integrating logs, threads, and code via multi-agent collaboration, Holmes reconstructs fragmented clues into causal chains, proving that \textbf{full-stack evidence alignment} is key to high-precision automated diagnosis.

\subsection{Performance Analysis}
\label{sec:performance}

\begin{figure*}[t]
  \centering
  \includegraphics[width=0.95\textwidth]{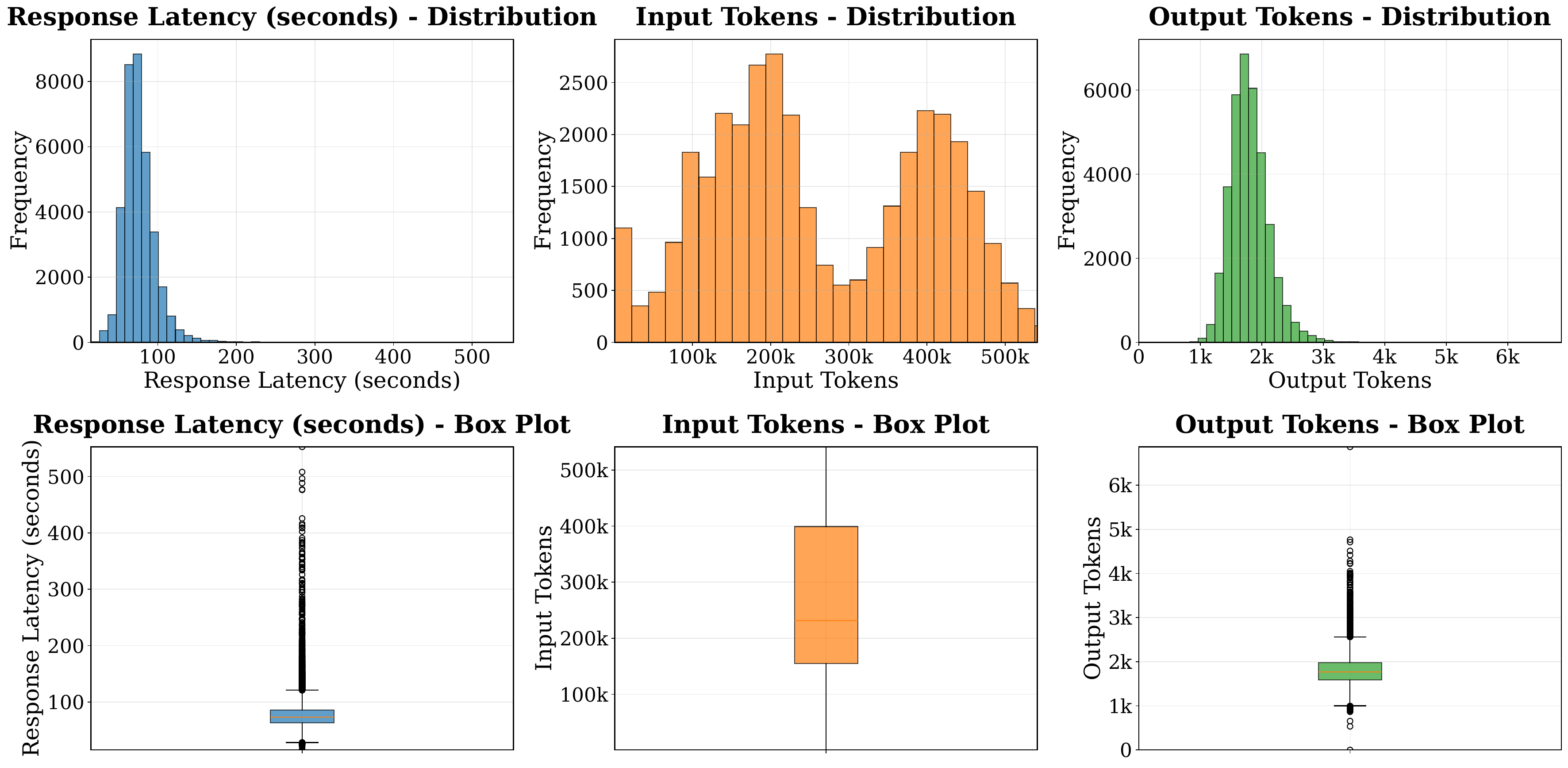}
  \caption{Statistical distribution of key performance metrics over 39,795 production runs.}
  \label{fig:distribution_analysis}
\end{figure*}

\begin{table*}[t]
  \centering
  \begin{minipage}{0.5\textwidth}
  \centering
  \small
  \setlength{\tabcolsep}{4pt}
  \renewcommand{\arraystretch}{1.08}
  \caption{Statistical distribution of performance metrics.}
  \label{tab:performance}
  {\arrayrulecolor{gray!30}\rowcolors{2}{gray!8}{white}%
  \begin{tabular*}{\linewidth}{@{\extracolsep{\fill}}>{\raggedright\arraybackslash}p{0.25\linewidth}rrr@{}}
  \rowcolor{gray!20}
  \textbf{Metric} & \textbf{Latency (s)} & \textbf{Input (Tokens)} & \textbf{Output (Tokens)} \\
  \midrule
  Mean & 77.02 & 267,194 & 1,803 \\
  Median & 73.44 & 231,575 & 1,768 \\
  Std. Dev. & 24.38 & 141,970 & 320 \\
  \bottomrule
  \end{tabular*}}
  \arrayrulecolor{black}
  \end{minipage}
  \hfill
  \begin{minipage}{0.45\textwidth}
  \centering
  \small
  \setlength{\tabcolsep}{3pt}
  \renewcommand{\arraystretch}{1.08}
  \caption{Pearson correlation coefficients.}
  \label{tab:correlation}
  {\arrayrulecolor{gray!30}\rowcolors{2}{gray!8}{white}%
  \begin{tabularx}{\linewidth}{@{}>{\raggedright\arraybackslash}X c@{}}
  \rowcolor{gray!20}
  \textbf{Correlation Pair} & \textbf{Coefficient} \\
  \midrule
  Latency $\leftrightarrow$ Output Tokens & 0.34 (Moderate positive) \\
  Latency $\leftrightarrow$ Input Tokens & -0.01 (None) \\
  Input Tokens $\leftrightarrow$ Output Tokens & -0.00 (None) \\
  \bottomrule
  \end{tabularx}}
  \arrayrulecolor{black}
  \end{minipage}
  \vspace{-10pt}
  \end{table*}

We analyzed Holmes' performance on 39,795 production runs (Table~\ref{tab:performance}, Figure~\ref{fig:distribution_analysis}).

\subsubsection{Statistical Summary}

\textbf{Latency (Time Cost):} With a median latency of 73.44s and mean of 77.02s (Std. Dev. 24.38s), Holmes reduces Time-to-Insight by >98\% compared to manual investigation (2--3 hours).
\textbf{Input Tokens (Cost):} Median input is stable at $\sim$232k tokens (Mean 267k, Std. Dev. 142k). The converted USD cost is $\sim$\$0.13 per session. This indicates high cost-effectiveness for large-scale deployment.
\textbf{Output Tokens (Summarization):} Output length is consistent (Mean 1,803, Median 1,768, Std. Dev. 320), summarizing contexts into $\sim$1,800 tokens.

\subsubsection{Correlation}
Pearson coefficients (Table~\ref{tab:correlation}) reveal no significant correlation between input tokens and latency (-0.01), confirming that the Map-Reduce architecture mitigates computational time costs of large contexts. Near-zero correlation between input and output tokens indicates output length depends on diagnostic complexity, not input volume.  A moderate positive correlation (0.34) exists between latency and output tokens.

\subsection{Improvement on the State of the Practice}

Beyond accuracy metrics, we evaluated Holmes' comprehensive impact on engineering efficiency, cost, and developer adoption.

\subsubsection{Efficiency Gains.}
\textbf{Manual Workflow:} Traditional diagnosis involves symbolication, manual git grep across 70M+ LOC, and log filtering. Internal telemetry shows this takes 2--3 hours per complex crash.
\textbf{Holmes Workflow:} The system generates a report in $\sim$77 seconds (Mean latency).
\textbf{Impact:} A 98\%+ reduction in ``Time to Insight,'' transforming developers from investigators to verifiers.

\subsubsection{Cost Efficiency Analysis.}
Running Holmes is orders of magnitude cheaper than human effort.
\textbf{Resource Consumption:} Due to the multi-agent parallel architecture, processing a single crash consumes an average of $\sim$267k input tokens (heavy log/code context) and $\sim$1.8k output tokens.
\textbf{ROI Calculation:} Comparing the salary cost of a senior engineer spending 2.5 hours (estimated $>\$70$) versus the API/GPU cost of a session ($\sim$\$0.13), the cost reduction approaches 99\%. This makes it economically viable to run Holmes on every new crash cluster.

\subsubsection{Developer Adoption \& Feedback.}
In a one-month pilot deployment with 45+ developers:
\textbf{Adoption:} The active adoption rate for complex crash tickets has reached 92\%.
\textbf{Feedback:} One senior developer noted: ``I used to dread assigning these Heisenbugs; now I check Holmes first.''
\textbf{Acceptance:} Developers accepted the AI's root cause analysis in 78\% of the cases without needing further manual log inspection.

\subsection{Case Study: Race Condition in MagicBrush Engine}

We present a representative case involving a race condition in the MagicBrush engine to illustrate Holmes's reasoning transparency. In this scenario, the application crashed with a \texttt{SIGBUS} error during module destruction. Holmes's agents collaborated to reconstruct the failure: the Log Miner and Thread Inspector identified a temporal conflict between the crashing thread (attempting to post a message) and a background thread (destroying the shared service), while the Code Explorer confirmed that the shared dictionary \texttt{bizToServiceMessages} was accessed without synchronization. The Synthesis Layer correctly diagnosed an atomicity violation where an object was deallocated by the background thread while being accessed by the main thread, and suggested adding an \texttt{NSRecursiveLock}, which resolved the issue.

%% file: sections/06-conclusion.tex
\section{Discussion}

\subsection{Limitations \& Failure Analysis}

\textbf{Privacy-Induced Information Entropy Reduction:} Strict privacy redaction (e.g., GDPR) removes sensitive data essential for diagnosis. In Case \#10, Holmes misdiagnosed \textit{corrupted input} as a generic memory error because the redacted message payload prevented tracing data dependencies.

\textbf{The Heisenbug Nature of Concurrency:} Race conditions are often invisible in static snapshots. In Case \#7, Holmes analyzed the victim thread in isolation, missing the temporal state inconsistency and misdiagnosing an assertion failure as a garbage pointer issue.

\textbf{Semantic Gap in Polyglot Runtimes:} In hybrid stacks (e.g., JNI, FFI), semantic fidelity can be lost across boundaries. In Case \#1, Holmes failed to interpret a high-level Swift exception and reverted to a low-level C++ error pattern (null pointer dereference).

\textbf{Non-Determinism of Memory Corruption:} Diagnosing memory safety issues often requires allocation history, but heavy instrumentation tools like AddressSanitizer or Malloc Stack Logging are unavailable in production due to excessive overhead. In Case \#12, Holmes observed only the crash consequence (invalid access) rather than the antecedent buffer overflow, causing a misdiagnosis.

\subsection{Lessons Learned}

\textbf{Architectural Determinism vs. Open-Ended Reasoning:} Ablation studies show that unstructured reasoning leads to attention drift (e.g., removing blame-frame anchoring drops accuracy by 6.1\%). The 3+1+1 architecture decomposes diagnosis into atomic, verifiable steps (Retrieval -> Exploration -> Synthesis), ensuring evidence chain integrity and reducing hallucinations compared to concatenating all artifacts into a single prompt.

\textbf{The Necessity of Semantic Filtering:} Feeding raw logs (e.g., 10k lines) triggers `lost-in-the-middle' issues. Semantic Map-Reduce is essential for RCA; filtering logs by relevance to the crash context improves the signal-to-noise ratio, enabling detection of subtle correlations.

\textbf{The Economic Value of Negative Feedback:} In industrial settings, false positives are costlier than false negatives. We explicitly instruct agents to report ``evidence not found'' rather than forcing a plausible but wrong explanation. This conservative bias builds developer trust and makes diagnoses actionable.

\subsection{Threats to Validity}

\textbf{Internal Validity.} The primary threat to internal validity lies in the potential subjectivity of the ground truth labels. Although we employed a rigorous cross-validation process with two senior developers to annotate the root causes and defect locations, manual labeling inherently carries a risk of human error or bias. To mitigate this, we resolved disagreements through discussion and excluded cases that were ambiguous.

\textbf{External Validity.} The results are based on data from a single large-scale application (WeChat iOS). While the dataset covers diverse modules (networking, storage, UI) and failure modes, the generalization to other platforms (e.g., Android, Server-side) remains to be fully verified. However, the core contribution—the 3-Stream reasoning paradigm—is platform-agnostic. By abstracting runtime artifacts into universal categories (Stack, Log, Thread) and decoupling retrieval from reasoning, Holmes can be adapted to other systems by simply integrating the respective symbolication toolchains (e.g., ProGuard/R8 mappings for Android, dSYM for iOS, DWARF for Linux), mitigating the risk of overfitting to a specific OS.

\textbf{Construct Validity.} We used Pass@1 accuracy and Time-to-Insight as primary metrics. While Pass@1 is a standard metric for fault localization, it may not fully capture the utility of the generated explanations or fix suggestions. A correct location with a misleading explanation could still hinder developers. To address this, we included a qualitative evaluation of the evidence chain and fix suggestions within our user study.

\textbf{Reliability Validity.} Large Language Models exhibit inherent non-determinism. Even with a fixed temperature (0.6), Holmes may produce slightly different reasoning paths for the same input across runs. To ensure robustness, we evaluated the system on a large-scale production dataset ($N=39,795$) and observed stable performance distributions, suggesting that the architectural constraints (e.g., blame-frame anchoring) effectively mitigate stochastic variations.

\section{Related Work}

We categorize the research landscape into three primary domains that address the challenges of automated diagnosis in large-scale systems.

\subsection{Evolution of Crash Analysis: From Text Similarity to Semantic Reasoning}

Industry tools such as Firebase Crashlytics and Sentry focus on crash clustering to handle large volumes of reports, and ReBucket \cite{ReBucket} provides a basis for call stack similarity metrics. In log analysis, LogAnomaly \cite{LogAnomaly} and LogPPT \cite{LogPPT} leverage sequence modeling and prompt-based few-shot learning to detect anomalies. More recent work, including LogLLM \cite{LogLLM} and FaithLog \cite{FaithLog}, further advances log understanding: LogLLM uses general-purpose LLMs to replace weaker pattern-matching pipelines, while FaithLog stresses diagnostic faithfulness, requiring the evidence chain to be verifiable against raw artifacts. However, most log-centric methods still treat logs as an isolated modality, lacking repository-scale, cross-modal reasoning over crash stacks and source code; consequently, they struggle to map runtime symptoms to specific defect-causing code locations. Holmes addresses this gap by combining an industrial Map-Reduce strategy with semantic filtering to integrate logs, stack traces, and code, bridging the semantic disconnect between runtime observations and navigation of repository-scale code.

\subsection{Repository-Scale Navigation: Static Indexes vs. Agentic Exploration}

Navigating massive codebases is a core challenge for SE agents. Approaches such as CodePlan \cite{CodePlan}, RepoGraph \cite{RepoGraph}, and LocAgent \cite{LocAgent} build fine-grained global indexes via static repository analysis to support more reliable context construction and cross-file reasoning. In this context, fine-grained global indexes refer to high-resolution semantic structures, such as complete Abstract Syntax Trees (ASTs), symbol cross-references (def-use / call-site) graphs, and pre-computed inter-procedural control-flow / call relations. While powerful, maintaining such dense indexes for production environments with millions of functions is prohibitively expensive due to the immense performance overhead and the volatility of industrial codebases. Recent empirical studies \cite{Wu2025} have attempted to enhance crash reports by utilizing LLMs to navigate codebases via such pre-built call graphs. This persistent reliance on static modeling reflects a broader limitation shared by traditional static analysis tools; for instance, Facebook Infer \cite{Infer} identifies potential defects based on code patterns but frequently suffers from severe scalability bottlenecks and a lack of runtime perspective. In ultra-large-scale repositories, the computational complexity of global inter-procedural analysis becomes intractable, and the absence of dynamic context often leads to high false-positive rates. Holmes addresses these scalability and precision bottlenecks by replacing dense semantic indexing with an agentic exploration paradigm supported by a shallow, lightweight function-to-file-path lookup table. By acting as a human investigator equipped with on-demand retrieval tools, Holmes autonomously navigates a 70M LOC repository, delegating logical cross-referencing to the agent's dynamic reasoning instead of a brittle pre-calculated graph.

\subsection{The Runtime Dependency Bottleneck: Sandboxed Execution vs. Post-mortem Reasoning}

Advancements in automated debugging generally fall into two categories: Workflow-based Fault Localization and Autonomous SE Agents. Traditional Spectrum-based Fault Localization (SBFL) \cite{Tarantula, SBFLAccuracy} relies on comparing execution traces from passing and failing tests. Modern LLM-based frameworks, such as Agentless \cite{Agentless} automate this navigation but still rely on a dynamic loop to verify hypotheses within a sandbox. Similarly, autonomous SE agents like SWE-agent \cite{SWEagent} and Openhands \cite{Openhands} follow an active coding paradigm where agents plan, edit, and verify fixes within a runnable environment. A common limitation of these approaches is their heavy dependency on reproducible runtime environments and failure-inducing tests, which are often unavailable in production crash scenarios. Holmes addresses this post-mortem challenge without re-running the application or relying on trial-and-error verification. Instead, Holmes employs an investigative reasoning paradigm to reconstruct the failure context solely from sparse, read-only artifacts (such as logs and stack snapshots) within a frozen, massive repository.

\section{Conclusion}

We have presented Holmes, a multi-agent framework that addresses the scalability and reproducibility challenges of post-mortem crash diagnosis in ultra-large-scale industrial systems. By synthesizing multimodal runtime signals—from high-level logs to low-level register states—Holmes bridges the semantic gap in mixed-source environments and efficiently navigates 70M+ LOC repositories without reproduction. Empirical evaluation on WeChat demonstrates 87.6\% fault localization accuracy and a 98\% reduction in investigation overhead, shifting the workflow from manual debugging to efficient verification. While validated on iOS, the underlying paradigm is platform-agnostic. Future work will focus on deploying lightweight, specialized models on local devices to enhance privacy and latency, as well as integrating automated reproduction and fix generation to achieve a closed-loop self-healing ecosystem.